\documentclass[11pt,fleqn]{article}

\usepackage[utf8]{inputenc}
\usepackage[T1]{fontenc}
\usepackage[a4paper,margin=1in]{geometry}
\usepackage{amsmath,amssymb}
\usepackage{graphicx}
\usepackage{booktabs}
\usepackage{makecell}
\usepackage{multirow}
\usepackage[table]{xcolor}
\usepackage{array}
\usepackage{tabularx}
\usepackage{float}
\usepackage[caption=false]{subfig}
\usepackage[protrusion=true,expansion=false]{microtype}
\usepackage[numbers,sort&compress]{natbib}
\usepackage{url}
\usepackage[hidelinks,hypertexnames=false]{hyperref}

\newcolumntype{Y}{>{\raggedright\arraybackslash}X}
\newcolumntype{Z}{>{\centering\arraybackslash}X}
\newcommand{\venue}[1]{\,\textsuperscript{\scriptsize #1}}
\title{Learnable Burst Quantization for Expressive and Efficient Spiking Neural Networks}
\author{%
Dewei Bai$^{1}$, Hongxiang Peng$^{1}$,
Hong Qu$^{1,*}$, and Dawen Xia$^{2}$\\[0.75em]
\small $^{1}$School of Computer Science and Engineering,\\
\small University of Electronic Science and Technology of China, Sichuan, China\\
\small $^{2}$Engineering Research Center of Micro-nano and Intelligent Manufacturing, Ministry of Education,\\
\small Kaili University, Guizhou, China\\[0.5em]
\small $^{*}$Corresponding author: \texttt{hongqu@uestc.edu.cn};
ORCID: \href{https://orcid.org/0000-0001-6114-3441}{0000-0001-6114-3441}%
}
\date{}
\hypersetup{
  pdftitle={Learnable Burst Quantization for Expressive and Efficient Spiking Neural Networks},
  pdfauthor={Dewei Bai, Hongxiang Peng, Jiajun Mei, Yang Reng, Hong Qu, Dawen Xia}
}

\begin{document}
\maketitle

\begin{abstract}
Binary spikes provide only two neuronal output states per timestep, limiting the response capacity of spiking neural networks (SNNs) under short simulation horizons. Burst neurons expand this response space, but their threshold spacing is typically fixed before training, leaving layer-specific burst resolution outside end-to-end optimization. We propose Learnable Burst Quantization (LBQ), which formulates burst emission as saturated uniform quantization with a positive, layer-wise learnable step. ReLSG-ET, a rectified-linear surrogate gradient with exponential tails, provides gradient support throughout and beyond the active burst range, thereby enabling joint optimization of synaptic weights and burst resolution. At inference, LBQ absorbs each learned step into downstream weights and decomposes integer burst levels into binary bit planes, accumulating only the non-zero planes. This changes the synaptic accumulation count for a level $S$ from $S$ to $\operatorname{popcount}(S)$. At two timesteps, LBQ achieves 97.45\% on CIFAR-10 and 82.82\% on CIFAR-100 with ResNet-20, and 73.67\% on ImageNet-1K with ResNet-34. On CIFAR-10, it comes within 0.07 percentage points of the 97.52\% ResNet-20 ANN reference; at $N_{\max}=5$, bit-plane execution reduces unary-equivalent synaptic accumulations by 40.52\% relative to unary execution. Controlled ablations isolate the benefits of learned quantization and ReLSG-ET, while layer-wise analyses reveal selective burst allocation across network depth. Results on CIFAR10-DVS and DVS128-Gesture extend the evidence to event-driven recognition. LBQ therefore couples adaptive burst resolution and accurate inference with an algebraically equivalent bit-sparse synaptic execution path.

\end{abstract}


\section{Introduction}
\label{sec:intro}

Accurate and efficient inference is a central objective for spiking neural networks (SNNs). Their sparse event communication supports accumulation-based synaptic computation on neuromorphic platforms such as Loihi~\cite{davies2018loihi} and Tianjic~\cite{pei2019towards}, making them attractive for energy-efficient perception~\cite{roy2019towards}. With short simulation horizons, however, each neuronal response must carry enough information to preserve recognition accuracy. Conventional binary neurons provide two output states per timestep, motivating richer discrete responses~\cite{guo_ternary_2024,zha2024energy}.

Burst and multi-level neurons address this representational constraint by emitting several response levels within one timestep. Multi-level spiking neurons~\cite{feng2022multilevel} and integer-valued LIF neurons~\cite{luo2024integer} demonstrate the value of this expanded response space. LM-HT~\cite{hao2024lmht} further connects equidistant multi-threshold firing with quantized ANN activations and learns temporal input mixing and membrane leakage. A complementary question remains: how should each layer set the spacing between its burst levels? When this spacing is fixed before training, the membrane-to-burst resolution cannot be optimized jointly with synaptic weights. The challenge is therefore not only to expand the response space, but also to optimize how each layer uses its finite burst levels for the recognition objective.

We propose Learnable Burst Quantization (LBQ), which formulates membrane-to-burst conversion as saturated uniform quantization with a positive step $\gamma^{(\ell)}$ learned separately for each layer. The step jointly sets the spacing between adjacent firing thresholds and the scale of the quantized output, allowing end-to-end training to optimize layer-specific burst resolution together with synaptic weights. To train this multi-level mapping, we introduce ReLSG-ET, a rectified-linear surrogate gradient with exponential tails. Its unit-gradient plateau spans the active burst range, while its exponential tails preserve learning signals in subthreshold and saturated regimes.

LBQ also connects the learned representation to bit-sparse synaptic execution. The quantization step is folded into downstream weights, preserving the scaled synaptic input while transmitting integer burst levels. Binary decomposition then represents each level as a set of non-zero bit planes, each triggering one synaptic accumulation of a power-of-two-scaled weight. This changes the accumulation count for a level $S$ from $S$ unary events to $\operatorname{popcount}(S)$ bit-plane events. The learned burst representation can therefore carry multiple response levels per timestep while exploiting sparsity within their binary encoding.

The results establish this combination of accuracy and computational structure. At two timesteps, LBQ reaches 97.45\% on CIFAR-10 and 82.82\% on CIFAR-100 with ResNet-20, improving over our architecture-matched LM-HT reproductions by 1.05 and 2.86 percentage points. On ImageNet-1K, ResNet-34 reaches 73.67\% at two timesteps. Controlled ablations identify the contribution of learned burst resolution, and layer-wise analyses show how the network allocates burst levels selectively. At $N_{\max}=5$, bit-plane execution reduces unary-equivalent synaptic accumulations by 40.52\% on CIFAR-10 while retaining near-ANN accuracy.

Our contributions are threefold:
\begin{itemize}
    \item We introduce LBQ, a layer-wise learnable membrane-to-burst quantizer, together with scale-consistent ReLSG-ET training to jointly optimize burst resolution and synaptic weights.
    \item We derive a synaptic-input-preserving transformation from scaled burst outputs to integer levels and bit-sparse synaptic accumulations, connecting learned quantization to an accumulation count governed by $\operatorname{popcount}(S)$ while separating bit-plane formation from backend-specific shifting.
    \item We demonstrate accurate few-timestep inference across static and event-driven benchmarks, supported by controlled ablations, layer-wise burst-allocation analyses, and quantified savings in synaptic accumulation counts.
\end{itemize}

\section{Related Work}
\label{sec:rel_work}

\subsection{SNN Training Paradigms}
Training methods for spiking neural networks (SNNs) primarily fall into two categories: ANN-to-SNN conversion and direct training.
ANN-to-SNN conversion leverages parameters from pre-trained ANNs to circumvent training difficulties. Techniques such as weight scaling, threshold balancing~\cite{dengoptimal}, and soft-reset mechanisms~\cite{dengoptimal} have successfully mapped analog ReLU activations to spike rates in deep architectures. Advanced optimization strategies, including quantization clip-floor-shift (QCFS)~\cite{buoptimal} and layer-wise calibration~\cite{liFreeLunchANN2021a}, align spiking rates with ANN activations to achieve high accuracy. However, this paradigm largely treats spikes as rate-based approximations of static values and often does not explicitly exploit fine-grained temporal dynamics, requiring long simulation windows to reduce quantization error.
Direct training utilizes surrogate gradients to address the non-differentiability of spike generation~\cite{neftci2019surrogate}. By introducing differentiable approximations for membrane potentials~\cite{lee2020enabling} and enforcing joint optimization of spatial and temporal dynamics (e.g., STBP~\cite{wu2018spatio}), these methods significantly reduce latency. Related temporal-error backpropagation methods also exploit spike-time errors for supervised learning in multilayer SNNs~\cite{luo2023stebp}. Our work adopts the direct training paradigm to fully leverage the temporal efficiency of SNNs. In addition to training paradigms, several works enhance SNN performance through architectural adaptations. Deng et al.~\cite{deng2024tensor} further introduced a tensor-decomposition-based attention module to exploit the tensor structure of spike representations. Such approaches focus on structural enrichment of spatial or spatiotemporal representations rather than modifying the spike coding mechanism itself.

\subsection{Burst Spiking Dynamics}
Binary spike coding constrains the information throughput of spiking neurons, which has motivated a series of studies aiming to mitigate quantization-induced information loss.
Early efforts have mainly focused on rectifying the membrane-potential distribution prior to binarization, including InfLoR-SNN~\cite{guo2023inflor}, IM-Loss~\cite{guo2022loss}, and RecDis-SNN~\cite{guo2022recdis}.
These methods alleviate information degradation at the distribution level, yet the spike output remains binary and thus intrinsically bounded by a 1-bit-per-timestep representational capacity.

Beyond distribution rectification, several works extend the discrete output space of spiking neurons.
Ternary Spikes~\cite{guo_ternary_2024} introduce signed spike states $\{-1,0,1\}$, providing a modest increase in expressiveness while still relying on a fixed and low-cardinality output space.
However, such limited extensions remain insufficient for modeling the wide dynamic ranges encountered in deep spiking networks.

More expressive spiking behaviors have been explored through multi-level or burst-based spiking mechanisms.
Multi-level spiking (MLF)~\cite{feng2022multilevel} enables neurons to emit multiple spikes within a single timestep by employing fixed, equidistant thresholds.
While MLF increases representational capacity, its rigid threshold design does not learn layer-specific threshold spacing. Peng et al.~\cite{peng2026bsvit} proposed BSViT, a burst spiking vision transformer that enhances visual representation learning through multi-level spike responses. 
Related integer-valued training paradigms, such as I-LIF~\cite{luo2024integer}, optimize networks using integer activations and decompose them into spike trains at inference.
Similarly, Spike-driven Transformer~\cite{yao2025scaling} adopts spiking approximations to facilitate large-scale SNN training, but primarily focuses on architectural scaling rather than learning adaptive spike quantization levels.

Hao et al.~\cite{hao2024lmht} proposed LM-HT, an equidistant multi-hierarchical threshold model that uses a learnable temporal-global information matrix and membrane-leakage parameters to regulate spike firing across timesteps. LM-HT connects multi-threshold firing to quantized ANN activations under uniform-current conditions, supports hybrid ANN--SNN training, and can be reparameterized into a single-threshold LIF model for deployment. LBQ is complementary in its target: instead of learning temporal current mixing, it directly optimizes the layer-wise spacing between adjacent membrane-to-burst levels and absorbs this learned step into downstream weights at inference.

From a hardware perspective, SNPU~\cite{kim2023snpu} introduced multi-level spike encoding, spike-train decomposition, and shift-and-accumulation hardware for processing compressed spike trains. LBQ connects these execution principles to learned membrane-to-burst resolution: quantization-step absorption preserves the scaled synaptic input, and bit-plane decomposition exposes sparsity within the resulting integer levels.

Fan et al.~\cite{fan_multisynaptic_2025} propose a multi-synaptic spiking neuron that enhances representational capacity by introducing parallel synaptic pathways with independent thresholds and dynamics. 
By distributing spiking responses across multiple synaptic channels, this model improves spatiotemporal information encoding without relying on high spiking rates. 
At a phenomenological level, such synaptic parallelism also increases the information carried by individual spike events, yielding effects partially analogous to burst spiking, but realized through structural expansion at the synaptic level rather than neuron-level multi-spike emission.

In parallel, burst spiking has been explored in ANN-to-SNN conversion frameworks~\cite{park2019fast, li_efficient_2022, lan2023efficient}, where burst spikes are leveraged to reduce conversion error and residual information loss. By permitting a neuron to emit multiple spikes within a single timestep, these methods effectively compress the simulation time required for accurate inference.
However, these approaches inherit the limitations of conversion-based pipelines and do not fully exploit burst dynamics during direct SNN training.

Many burst spiking methods implicitly implement uniform quantization through fixed threshold spacings, leaving the discretization granularity outside joint optimization with synaptic weights. LM-HT introduces learnability into the temporal input and leakage dynamics of an equidistant multi-threshold model~\cite{hao2024lmht}, but does not formulate layer-specific threshold spacing as the primary learned variable. Although learnable step-size quantization has also been explored in low-precision ANNs, such as LSQ~\cite{esser2020learned}, its objective is to quantize weights and activations for low-precision numerical computation. In contrast, LBQ learns a membrane-to-burst quantization step for spike-event generation, where the integer quantization level directly represents burst multiplicity and affects downstream synaptic accumulation.

\subsection{Surrogate Gradients for Burst Spiking}

The non-differentiability of spike generation is one of the central challenges in directly training spiking neural networks (SNNs). Surrogate-gradient learning addresses this issue by replacing the zero derivative of the spike function with a smooth approximation during backpropagation while preserving discrete spikes in the forward pass \cite{neftci2019surrogate}. Most existing surrogate functions were developed for binary spiking neurons with a single firing threshold. Representative examples include rectangular and sigmoid derivative approximations in STBP \cite{wu2018spatio}, piecewise-linear approximations for binary-valued neurons \cite{esser2016convolutional}, arctangent variants \cite{fangDeepResidualLearning2021,fang2021incorporatinglearnablemembranetime}, and SuperSpike \cite{zenke2018superspike}. Although these functions differ in shape, they share a common design principle: concentrating gradients around a single threshold-crossing event. This formulation has proven effective for binary SNNs and remains a dominant training paradigm in directly trained SNN architectures \cite{zenke2021remarkable}. Recent work has also combined adaptive neuronal thresholds with learnable surrogate-gradient scaling to improve the optimization of deep residual SNNs~\cite{ai2025cross}; however, the spike-generation process remains binary and centered on a single effective firing threshold. 

Burst and multi-level spiking neurons require a different surrogate-gradient design because the membrane potential may span multiple emission intervals within a single timestep. Multi-level firing neurons \cite{feng2022multilevel} and integer-valued LIF neurons \cite{luo2024integer} enlarge the spike output space beyond binary events, but this also changes the backward optimization problem: gradients must remain informative across several burst levels rather than only near one threshold. In this setting, conventional single-threshold surrogates such as sigmoid and arctangent derivatives can decay rapidly as the membrane potential moves into higher burst intervals, which may weaken the optimization of high-level burst responses. Heidarian et al.~\cite{heidarian2024effective} introduced a temporal feedback backpropagation mechanism for effective multispike learning, demonstrating the potential of richer spike timing and spike-count representations in supervised SNN training.
Rectangular or box-shaped surrogates alleviate this issue by maintaining an approximately constant gradient over a wider valid burst-emission interval \cite{luo2024integer}. Nevertheless, existing box-shaped surrogates introduce hard cutoffs at the interval boundaries, yielding zero gradients for both sub-threshold and saturated membrane states.

As a result, neurons that fall outside the valid burst range may receive little or no corrective learning signal. 
LBQ addresses this limitation through the proposed ReLSG-ET surrogate, which preserves the constant-gradient plateau within the burst-emission interval while introducing exponentially decaying tails outside the valid range.

\begin{figure*}[t]
    \centering
    \includegraphics[width=0.88\linewidth]{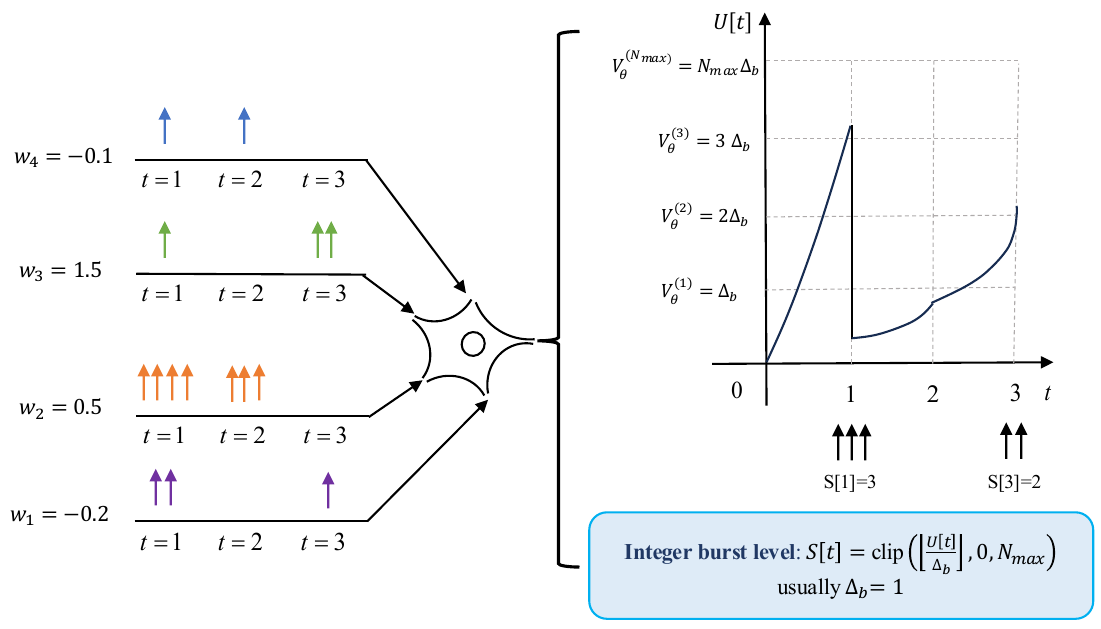}
   \caption{\textbf{Fixed-step burst spiking neuron dynamics.}
Spike inputs from multiple channels are integrated into the membrane potential $U[t]$, and uniformly spaced thresholds
${\Delta_b,2\Delta_b,\ldots,N_{\max}\Delta_b}$ define a fixed quantization ladder.
The emitted burst level $S[t]$ is determined by the number of crossed threshold intervals, while reset is triggered by any non-zero burst level.}
    \label{fig:fixed_burst}
\end{figure*}

\section{Preliminaries}
\label{sec:prelim}

\subsection{Leaky Integrate-and-Fire Neuron}
\label{subsec:lif_neuron}

Although the proposed learnable quantization mechanism can be incorporated into different membrane-potential-based spiking neurons, we instantiate it using the LIF neuron in this work because of its simplicity and widespread adoption in modern SNNs~\cite{buoptimal, liFreeLunchANN2021a, dengoptimal}.

In standard LIF neurons, the membrane potential $U[t]$ evolves over time according to:

\begin{equation}
\label{eq:lif}
U[t]
=
\underbrace{\beta U[t-1]}_{\text{decay}}
+
\underbrace{W X[t]}_{\text{input}}
-
\underbrace{\alpha V_{\theta}S[t-1]}_{\text{subtractive soft reset}},
\end{equation}
and
\begin{equation}
\label{eq:bispiking}
  S[t] =
  \begin{cases}
  1 & \text{if } U[t] > V_{\theta}, \\[6pt]
  0 & \text{otherwise.}
  \end{cases}
\end{equation}

Here, $U[t]$ denotes the membrane potential at time $t$, $V_{\theta}$ is the firing threshold, $\beta$ represents the decay rate, $\alpha\in[0,1]$ controls the subtractive reset strength, $W$ is the learnable synaptic weight matrix, and $X[t]$ is the input at time $t$. When $\alpha=1$, a spike at the preceding timestep subtracts one threshold from the membrane potential while preserving both the residual potential and the current input $WX[t]$, which is the conventional soft-reset operation. A binary spike $S[t] = 1$ is emitted when the membrane potential exceeds the threshold. This binary LIF neuron remains one of the most widely used spiking models due to its computational efficiency and biological plausibility.

\subsection{Burst Spiking Neuron}
\label{subsec:burst_spiking_neuron}

Burst coding is an important neural coding scheme in SNNs, in which a neuron can transmit information through a burst of spikes rather than a single binary event~\cite{li_efficient_2022}.
Compared with binary spike coding, burst coding can improve information transmission capacity within a few timesteps, making it attractive for low-latency SNN inference.
Recent deep SNN studies have also explored multi-level or integer-valued spike states to reduce the information loss caused by binary thresholding~\cite{luo2024integer}.

As illustrated in Fig.~\ref{fig:fixed_burst}, a burst level denotes the number of thresholds crossed by the membrane potential at a timestep.
Let
\(
\{V_{\theta}^{(i)}\}_{i=1}^{N_{\max}}
\)
denote an ordered set of burst thresholds satisfying
\(
V_{\theta}^{(1)}
<
V_{\theta}^{(2)}
<
\cdots
<
V_{\theta}^{(N_{\max})}
\).
A generic burst spiking neuron can be written as
\begin{equation}
\label{eq:burst_count}
S[t]
=
\sum_{i=1}^{N_{\max}}
\mathbb{I}
\left[
U[t]\ge V_{\theta}^{(i)}
\right],
\end{equation}
where
\(
S[t]\in\{0,1,\dots,N_{\max}\}
\)
denotes the emitted burst level.
This formulation indicates that the burst level is determined by the number of thresholds crossed by the membrane potential.

When the thresholds are uniformly spaced, the above multi-threshold formulation can be interpreted as a fixed-step quantization of the membrane potential:
\begin{equation}
\label{eq:fixed_burst_quantizer}
S[t]
=
\operatorname{clip}
\left(
\left\lfloor
\frac{U[t]}
{\Delta_b}
\right\rfloor,
0,
N_{\max}
\right),
\end{equation}
where
\(
\Delta_b
\)
denotes the fixed threshold interval.
This quantization view makes the role of burst firing explicit:
\(N_{\max}\)
controls the maximum burst level, while
\(\Delta_b\)
controls the discretization granularity of the membrane potential.

\section{Methodology}
\label{sec:method}

\begin{figure*}[t]
    \centering
    \includegraphics[width=0.98\linewidth]{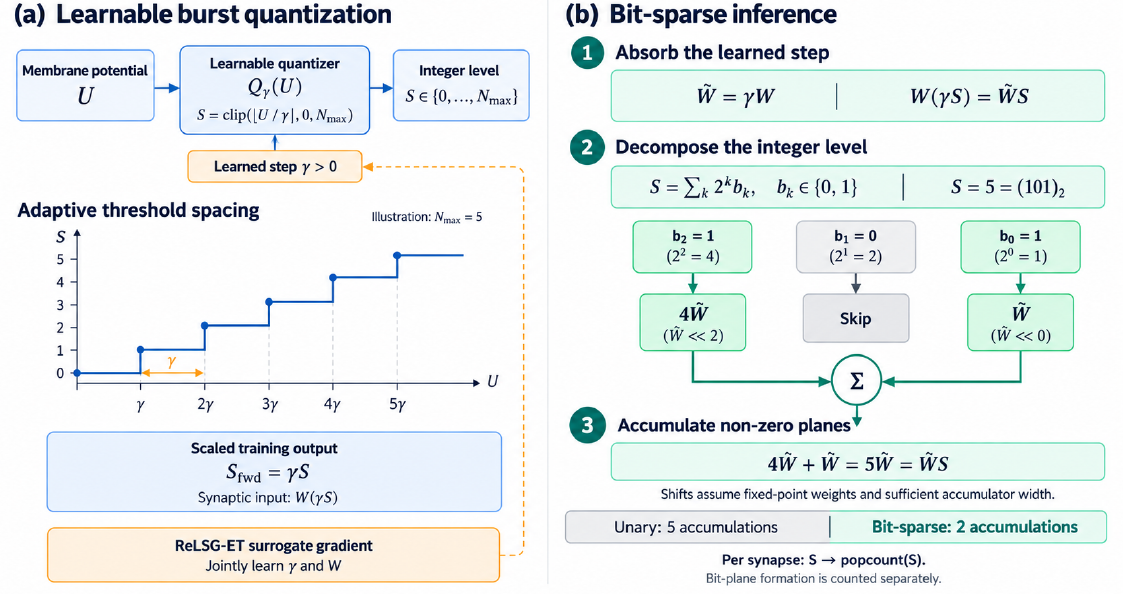}
    \caption{
\textbf{Overview of Learnable Burst Quantization (LBQ).}
(a) A positive layer-wise step $\gamma^{(\ell)}$ controls the threshold spacing that maps membrane potential $U^{(\ell)}[t]$ to integer burst level $S^{(\ell)}[t]$. The scaled training output is $S_{\operatorname{fwd}}^{(\ell)}[t]=\gamma^{(\ell)}S^{(\ell)}[t]$, and ReLSG-ET supports joint learning of the step and synaptic weights.
(b) Inference absorbs the output scale into downstream weights, $\widehat{W}^{(\ell+1)}=\gamma^{(\ell)}W^{(\ell+1)}$, and accumulates only non-zero bit planes. For $S=5=(101)_2$, two power-of-two-scaled weight terms yield $4\widehat{W}+\widehat{W}=5\widehat{W}$, replacing five unary accumulations per synapse. The reported accumulation count excludes bit-plane formation and explicit runtime shifts: formation is counted separately, while a target-specific shift term is added only for backends that issue separate shift instructions. Layer and time indices are omitted in the diagram, and $W$ denotes the downstream weight.
}
    \label{fig:lbq_overview}
\end{figure*}

Fig.~\ref{fig:lbq_overview} illustrates the training-time quantization and inference-time bit-sparse execution pipeline of Learnable Burst Quantization (LBQ). During training, LBQ learns a layer-wise quantization step $\gamma^{(\ell)}$ to map membrane potentials into integer burst levels and represent them as scaled burst outputs for downstream synaptic computation. During inference, the output scale is absorbed into downstream weights while the learned step continues to determine the neuronal quantization thresholds. Each emitted integer level is then decomposed into bit planes, and only non-zero planes trigger synaptic accumulation of the corresponding power-of-two-scaled weights, preserving the scaled synaptic input. The following subsections define the LBQ formulation, introduce the ReLSG-ET surrogate gradient, and derive this exact inference transformation.

\subsection{Learnable Burst Quantization}
\label{subsec:lbq_model}

Existing burst or integer-valued spiking neurons usually rely on manually specified or implicitly fixed quantization intervals. Consequently, the discretization granularity of each layer is prescribed before training rather than optimized with synaptic weights, limiting how the network can allocate its finite burst levels for the recognition objective.

To address this limitation, we propose Learnable Burst Quantization (LBQ).
As illustrated in Fig.~\ref{fig:lbq_overview}(a), LBQ preserves the integer-valued burst representation while replacing the fixed interval $\Delta_b$ in Eq.~\eqref{eq:fixed_burst_quantizer} with a positive layer-wise learnable quantization step $\gamma^{(\ell)}$.
In this work, LBQ is instantiated with LIF membrane dynamics.
To ensure positivity, we parameterize $\gamma^{(\ell)}$ as
$\gamma^{(\ell)}=\operatorname{Softplus}\!\left(\rho^{(\ell)}\right)+\epsilon$,
where $\rho^{(\ell)}$ is an unconstrained learnable parameter and $\epsilon$ is a small constant for numerical stability.

For layer $\ell$, let $U^{(\ell)}[t]$ denote the membrane potential before spike emission and reset, and let $V^{(\ell)}[t]$ denote the post-reset state. LBQ maps $U^{(\ell)}[t]$ to a scaled quantized burst output:
\begin{equation}
\label{eq:lbq_fire}
S_{\operatorname{fwd}}^{(\ell)}[t]
=
\gamma^{(\ell)} S^{(\ell)}[t]
=
\gamma^{(\ell)}
\operatorname{clip}
\left(
\left\lfloor
\frac{U^{(\ell)}[t]}{\gamma^{(\ell)}}
\right\rfloor,
0,
N_{\max}
\right),
\end{equation}
where the integer burst level is defined as
\begin{equation}
\label{eq:lbq_level}
S^{(\ell)}[t]
=
\operatorname{clip}
\left(
\left\lfloor
\frac{U^{(\ell)}[t]}{\gamma^{(\ell)}}
\right\rfloor,
0,
N_{\max}
\right)
\in\{0,1,\dots,N_{\max}\}.
\end{equation}
Here, $N_{\max}$ controls the maximum burst capacity at each timestep.
In this formulation, $S^{(\ell)}[t]$ determines the event multiplicity, while $\gamma^{(\ell)}$ specifies the layer-wise voltage resolution.
A smaller $\gamma^{(\ell)}$ yields finer membrane-potential discretization, whereas a larger one produces coarser burst levels. Learning this step adapts the firing-threshold spacing and output scale jointly.

Given the scaled training output from the previous layer, an LBQ-equipped LIF neuron first decays the previous post-reset state and integrates the incoming current:
\begin{equation}
\label{eq:lif_burst_integrate}
U^{(\ell)}[t]
=
\beta^{(\ell)}V^{(\ell)}[t-1]
+
W^{(\ell)}S_{\mathrm{fwd}}^{(\ell-1)}[t].
\end{equation}
It then emits the burst defined in Eqs.~\eqref{eq:lbq_fire}--\eqref{eq:lbq_level} and applies a subtractive soft reset:
\begin{equation}
\label{eq:lif_burst_reset}
V^{(\ell)}[t]
=
U^{(\ell)}[t]
-
\alpha^{(\ell)}\gamma^{(\ell)}
\mathbb{I}
\!\left[
S^{(\ell)}[t]>0
\right]
.
\end{equation}

Here, $\beta^{(\ell)}$ and $\alpha^{(\ell)}$ denote the membrane decay factor and the soft-reset factor, respectively, and the post-reset state is initialized to zero before processing each input sequence.
A non-zero response at the current timestep triggers subtraction of $\alpha^{(\ell)}\gamma^{(\ell)}$. The resulting post-reset state is carried to the next timestep, where it is multiplied by $\beta^{(\ell)}$ before the new synaptic input is added. The integer burst magnitude controls downstream synaptic accumulation, while its non-zero indicator controls reset. We optimize the floor-quantized mapping with the scale-consistent surrogate rule below.

\subsection{Scale-Consistent Surrogate Training}
\label{subsec:surrogate}

Since the quantized burst mapping is piecewise constant with respect to the membrane potential, it is not directly amenable to gradient-based optimization.
To preserve the coupling between forward discretization and backward gradient propagation, LBQ defines the surrogate on the normalized membrane variable $r^{(\ell)}[t] = \frac{U^{(\ell)}[t]}{\gamma^{(\ell)}}$ and introduces the Rectified-Linear Surrogate Gradient with Exponential Tails (ReLSG-ET) through the continuous surrogate level function:
\begin{equation}
\label{eq:surrogate_level}
B(r)
=
\begin{cases}
1+\lambda_{\mathrm{tail}}(e^{r-1}-1), & r<1,\\[3pt]
r, & 1\le r\le N_{\max},\\[3pt]
N_{\max}+\lambda_{\mathrm{tail}}(1-e^{N_{\max}-r}),
& r>N_{\max}.
\end{cases}
\end{equation}
where $\lambda_{\mathrm{tail}}\in(0,1)$ is a learnable tail coefficient that controls the strength of the exponential tails. We parameterize it as
$\lambda_{\mathrm{tail}}=\operatorname{sigmoid}(\eta_{\mathrm{tail}})$
to keep it within $(0,1)$ during training.

\begin{figure*}[t]
\centering
\includegraphics[width=0.65\linewidth]{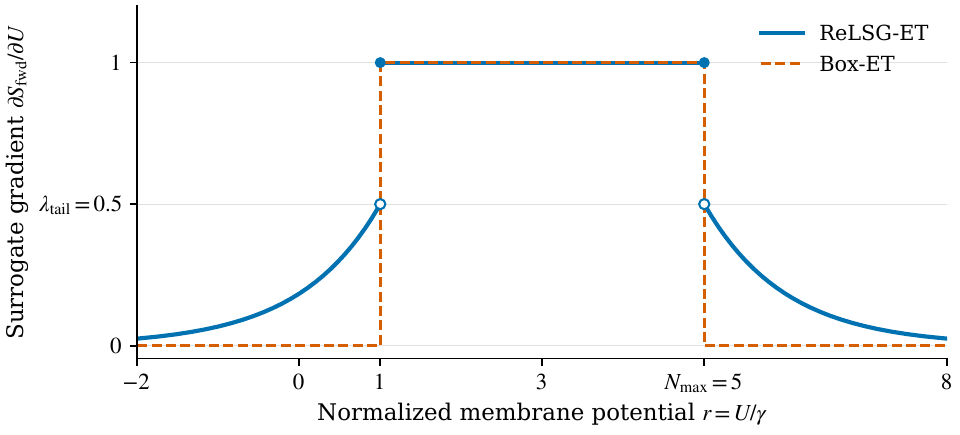}
\caption{Surrogate gradients as a function of normalized membrane potential $r=U/\gamma$. ReLSG-ET provides a unit-gradient plateau over $[1,N_{\max}]$ and exponential tails with strength $\lambda_{\mathrm{tail}}$. The illustrative curves use $N_{\max}=5$ and $\lambda_{\mathrm{tail}}=0.5$; Box-ET denotes the box surrogate used in the ablations.}
\label{fig:surrogate}
\end{figure*}

Figure~\ref{fig:surrogate} illustrates the gradient support of ReLSG-ET. The unit-gradient plateau covers multiple burst levels, while the exponential tails supply non-zero learning signals outside the active interval. The tail coefficient controls their strength independently of the plateau. Writing $r=r^{(\ell)}[t]$, we implement the straight-through estimator:

\begin{equation}
\label{eq:ste_level}
S_{\operatorname{fwd}}^{(\ell)}[t]
=
\gamma^{(\ell)}
\left[
B(r)
+
\operatorname{sg}
\left(
S^{(\ell)}[t]
-
B(r)
\right)
\right],
\end{equation}
where $\operatorname{sg}(\cdot)$ denotes the stop-gradient operator. In the forward pass, Eq.~\eqref{eq:ste_level} exactly recovers the scaled training output $\gamma^{(\ell)}S^{(\ell)}[t]$. In the backward pass, the derivative is determined by $B(r)$, yielding
\begin{equation}
\label{eq:surrogate_grad}
\frac{
\partial S_{\operatorname{fwd}}^{(\ell)}[t]
}{
\partial U^{(\ell)}[t]
}
\approx
\begin{cases}
\lambda_{\mathrm{tail}}e^{r-1},
& r<1,\\[3pt]
1,
& 1\le r\le N_{\max},\\[3pt]
\lambda_{\mathrm{tail}}e^{N_{\max}-r},
& r>N_{\max}.
\end{cases}
\end{equation}

The same scale-consistent formulation also enables direct optimization of the quantization step. In the active range, differentiating Eq.~\eqref{eq:ste_level} with respect to $\gamma^{(\ell)}$ gives
\begin{equation}
\label{eq:scale_grad_middle}
\frac{\partial S_{\operatorname{fwd}}^{(\ell)}[t]}
{\partial \gamma^{(\ell)}}
\approx
S^{(\ell)}[t]-r^{(\ell)}[t],
\qquad
1\le r^{(\ell)}[t]\le N_{\max}.
\end{equation}
Thus, $\gamma^{(\ell)}$ receives a learning signal from the discrepancy between the integer burst level and the normalized membrane potential. This differs from amplitude-rescaling strategies~\cite{guo_ternary_2024, guo2022loss}: here, $\gamma^{(\ell)}$ is not merely a post-spike gain, but the quantization step that simultaneously determines the forward burst resolution and the backward gradient support.

\subsection{Inference-Time Step Absorption and Bit-Sparse Execution}
\label{subsec:quantization_step_absorption}

As illustrated in Fig.~\ref{fig:lbq_overview}(b), the scaled training output does not require real-valued spike transmission at inference. Since the learned quantization step $\gamma^{(\ell)}$ enters linearly into the synaptic input to the next layer, it can be absorbed into the downstream weights:
\begin{equation}
\label{eq:absorb}
\begin{aligned}
W^{(\ell+1)}S_{\operatorname{fwd}}^{(\ell)}[t]
&=
W^{(\ell+1)}\gamma^{(\ell)}S^{(\ell)}[t]
=
\widehat{W}^{(\ell+1)}S^{(\ell)}[t],
\\
\widehat{W}^{(\ell+1)}
&=
\gamma^{(\ell)}W^{(\ell+1)} .
\end{aligned}
\end{equation}
Therefore, the deployed neuron emits only the integer burst level $S^{(\ell)}[t]\in\{0,\dots,N_{\max}\}$, while the effect of the learned membrane-to-burst quantization step is retained by the absorbed synaptic weights.

Let $K^{(\ell)}=\lceil\log_2(N_{\max}+1)\rceil$ denote the number of bit planes required by the configured level range. Each integer burst level has the exact binary decomposition
\begin{equation}
\label{eq:bitplane_decomp}
S^{(\ell)}[t]
=
\sum_{k=0}^{K^{(\ell)}-1}2^k b_k^{(\ell)}[t],
\qquad b_k^{(\ell)}[t]\in\{0,1\}.
\end{equation}
Substituting Eq.~\eqref{eq:bitplane_decomp} into the absorbed synaptic product yields
\begin{equation}
\label{eq:bit_sparse_product}
\widehat{W}^{(\ell+1)}S^{(\ell)}[t]
=
\sum_{k:\,b_k^{(\ell)}[t]=1}
2^k\widehat{W}^{(\ell+1)}.
\end{equation}
Equations~\eqref{eq:absorb}--\eqref{eq:bit_sparse_product} preserve the scaled synaptic input algebraically without prescribing a particular backend realization. On fixed-point hardware, the factor $2^k$ may be realized using pre-shifted weights, a fused shift--accumulate datapath, or an explicit runtime shift; zero bit planes are skipped in all three cases.

For the non-negative LBQ levels indexed by $j$, with downstream synaptic fan-out $F_j$, we distinguish two accumulation counts. Unary execution requires
\begin{equation}
\label{eq:unary_acs}
C_{\mathrm{AC}}^{\mathrm{unary}}
=
\sum_j F_j S_j,
\end{equation}
which treats a level $S_j$ as $S_j$ unit-magnitude synaptic accumulations. Bit-sparse execution instead requires
\begin{equation}
\label{eq:bit_sparse_acs}
\begin{aligned}
C_{\mathrm{AC}}^{\mathrm{BS}}
&=
\sum_j F_j\operatorname{popcount}(S_j),
\\
C_{\mathrm{BPF}}
&=
\sum_j\operatorname{popcount}(S_j),
\end{aligned}
\end{equation}
so the accumulation count is proportional to the number of non-zero bits rather than the emitted burst magnitude. Here $C_{\mathrm{AC}}^{\mathrm{unary}}$ and $C_{\mathrm{AC}}^{\mathrm{BS}}$ count fan-out-weighted accumulations, whereas $C_{\mathrm{BPF}}$ assigns one abstract 32-bit integer ALU operation per exposed non-zero bit before fan-out. None includes an explicit runtime weight shift. For Binary/Ternary, the corresponding unit-event count is \mbox{$\sum_jF_j\mathbf{1}[S_j\neq0]$}; the sign selects addition or subtraction.

For a backend that issues power-of-two scaling separately, the fan-out-weighted shift count, with $K_j=K^{(\ell(j))}$, is
\begin{equation}
\label{eq:explicit_shifts}
C_{\mathrm{shift}}^{\mathrm{syn}}
=
\sum_j F_j\sum_{k=1}^{K_j-1}b_{j,k},
\end{equation}
where the $k=0$ plane requires no shift. This explicit instruction count is zero when shifted weights are precomputed or shifting is fused into the accumulation datapath.

For tensors connected to multiple downstream synaptic transformations, the same absorption is applied to each corresponding weight kernel. For additive merging paths, the involved operands are kept in a common scale space before summation. These operations do not alter the neuron model itself but ensure that the algebraic equivalence in Eq.~\eqref{eq:absorb} is preserved in network architectures with branching or residual connections. The execution path requires zero-bit skipping and sparse accumulation; power-of-two scaling may be precomputed, fused, or issued explicitly according to the target backend.

\section{Experiments}
\label{sec:exp}

We evaluate three aspects of LBQ: recognition accuracy at short simulation horizons, the contribution of learned burst resolution, and the arithmetic savings enabled by bit-plane execution. Static benchmarks include CIFAR-10, CIFAR-100~\cite{krizhevsky2009cifar10}, and ImageNet-1K~\cite{93443fe0aca44aa4b4f773517d12ee30}; event-driven benchmarks include CIFAR10-DVS~\cite{liCIFAR10DVSEventStreamDataset2017} and DVS128-Gesture~\cite{8100264}. We use ResNet-18, ResNet-19, ResNet-20, and ResNet-34 architectures following standard SNN designs~\cite{fangDeepResidualLearning2021,zhengGoingDeeperDirectlytrained2021,9556508}.

\subsection{Experimental Setup}
The ImageNet-1K and CIFAR-10/100 training protocols follow the public LM-HT implementation~\cite{hao2024lmht}, with LBQ-specific timestep and maximum-burst settings.\footnote{LM-HT implementation: \url{https://github.com/hzc1208/LMHT_SNN/}} We use SGD with momentum 0.9 and Nesterov acceleration on these datasets. Data augmentation includes random cropping, horizontal flipping, AutoAugment~\cite{cubuk2019autoaugment}, and dataset-specific Cutout~\cite{devries2017improved} or mixing augmentations, as summarized in Table~\ref{tab:set}. Event-driven models use AdamW. All experiments run on a workstation with an AMD EPYC 7282 16-core CPU and eight NVIDIA A100 GPUs.

For architecture-matched comparison, we reproduce LM-HT on CIFAR-10, CIFAR-100, and CIFAR10-DVS using ResNet-20; these results are marked with $\dagger$ in Tables~\ref{tab:cifar} and~\ref{tab:dvs_eval}. The ablation and arithmetic-cost comparisons use identical training settings within our implementation.

\begin{table*}[t]
\centering
\caption{Training hyperparameters for image and event-based benchmarks.}
\label{tab:set}
\footnotesize
\setlength{\tabcolsep}{3pt}
\renewcommand{\arraystretch}{1.06}
\begin{tabularx}{\textwidth}{@{}lZZZZ@{}}
\toprule
Hyperparameter & ImageNet-1K & CIFAR-10/100 & CIFAR10-DVS & DVS128-Gesture \\
\midrule
Max spikes & 5 & 5 & 5 & 5 \\
Timestep & $1/2$ & $1/2/4$ & 10 & 16 \\
Epochs & 320 & 300 & 300 & 300 \\
Resolution & $224\times224$ & $32\times32$ & $128\times128$ & $128\times128$ \\
Batch size & 64 & 64 & 64 & 64 \\
Optimizer & SGD & SGD & AdamW & AdamW \\
Base learning rate & $1\times10^{-1}$ & $2.5\times10^{-2}$ & $1\times10^{-3}$ & $1\times10^{-3}$ \\
Learning rate decay & Cosine & Cosine & Cosine & Cosine \\
Warmup epochs & None & None & 10 & 10 \\
Weight decay & 0 & $5\times10^{-4}$ & 0.06 & 0.06 \\
AutoAugment & Yes & Yes & -- & -- \\
Cutout length & -- & $16/12$ (C10/C100) & -- & -- \\
Mixup & 0.8 & None & 0.5 & 0.5 \\
Cutmix & 1.0 & None & 0.0 & 0.0 \\
Mixup-off epoch & None & None & None & None \\
Label smoothing & 0.1 & 0.0 & 0.1 & 0.1 \\
\bottomrule
\end{tabularx}
\end{table*}

\subsection{Performance on Static Datasets}
 
\subsubsection{CIFAR}
LBQ achieves 97.45\% on CIFAR-10 and 82.82\% on CIFAR-100 with ResNet-20 at two timesteps (Table~\ref{tab:cifar}). With a single timestep, it already reaches 97.28\% and 82.38\%, respectively, demonstrating that learned burst resolution supports accurate recognition within a short simulation horizon.

The architecture-matched LM-HT reproductions provide a direct comparison at each timestep. LBQ improves CIFAR-10 accuracy by 1.35, 1.05, and 1.36 percentage points at $T=1,2,4$, respectively; the corresponding CIFAR-100 gains are 3.64, 2.86, and 2.04 points. The CIFAR-100 result is particularly strong at $T=1$, where LBQ reaches 82.38\% against 78.74\% for LM-HT. Across all three horizons, learning the membrane-to-burst mapping yields higher accuracy within the same architecture and timestep budget. Section~\ref{subsec:ablation} examines the contribution of the quantization step under controlled settings.

\begin{table*}[t]
  \centering
  \caption{Classification accuracy (\%) on CIFAR-10 and CIFAR-100 at different simulation timesteps.}
  \label{tab:cifar}
  \footnotesize
  \setlength{\tabcolsep}{3pt}
  \renewcommand{\arraystretch}{1.04}
  \begin{tabularx}{\textwidth}{@{}lYlccc@{}}
  \toprule
  Dataset & Method & Architecture & T=1 & T=2 & T=4 \\
  \midrule

  \textbf{CIFAR-10} & \multicolumn{5}{@{}l}{\textit{ANN-to-SNN conversion}} \\
  & RMP-SNN~\cite{hanRmpsnnResidualMembrane2020}\venue{CVPR-2020}
  & ResNet-20 & \multicolumn{3}{c}{91.36 (T = 2048)} \\
  & Burst+LIPooling~\cite{li_efficient_2022}\venue{IJCAI-2022}
  & ResNet-20 & \multicolumn{3}{c}{96.49 (T = 64)} \\

  \cmidrule(lr){2-6}
  & \multicolumn{5}{@{}l}{\textit{Direct SNN training}} \\
  & GLIF~\cite{yao2023glifunifiedgatedleaky}\venue{NeurIPS-2022}
  & ResNet-19 & -- & 94.44 & 94.85 \\
  & TET~\cite{deng2022temporalefficienttrainingspiking}\venue{ICLR-2022}
  & ResNet-19 & -- & 94.16 & 94.44 \\
  & RecDis-SNN~\cite{guo2022recdis}\venue{CVPR-2022}
  & ResNet-19 & -- & 93.64 & 95.53 \\
  & MLF~\cite{feng2022multilevel}\venue{IJCAI-2022}
  & ResNet-19 & -- & -- & 94.25 \\
  & LT-SNN~\cite{hasssan2024spiking}\venue{IJCNN-2024}
  & ResNet-19 & -- & 94.19 & -- \\
  & Ternary Spike~\cite{guo_ternary_2024}\venue{AAAI-2024}
  & ResNet-20 & -- & 94.48 & 94.96 \\
  & DA-LIF~\cite{10888909}\venue{ICASSP-2025}
  & ResNet-20 & 92.89 & 93.65 & 94.16 \\
  & FSTA-SNN~\cite{yu2025fstasnnfrequencybasedspatialtemporalattentionmodule}\venue{AAAI-2025}
  & ResNet-20 & 93.01 & 94.18 & 94.72 \\
  & STAA-SNN~\cite{zhang2025staasnnspatialtemporalattentionaggregator}\venue{CVPR-2025}
  & ResNet-20 & 93.08 & 94.35 & 95.03 \\
  & LM-HT ($L=2$)$^{\dagger}$~\cite{hao2024lmht}\venue{NeurIPS-2024}
  & ResNet-20 & 95.93 & 96.40 & 96.13 \\
  \cmidrule(lr){2-6}
  & LBQ
  & ResNet-20 & \textbf{97.28} & \textbf{97.45} & \textbf{97.49} \\

  \midrule

  \textbf{CIFAR-100} & \multicolumn{5}{@{}l}{\textit{ANN-to-SNN conversion}} \\
  & RMP-SNN~\cite{hanRmpsnnResidualMembrane2020}\venue{CVPR-2020}
  & ResNet-20 & \multicolumn{3}{c}{67.82 (T = 2048)} \\
  & Burst+LIPooling~\cite{li_efficient_2022}\venue{IJCAI-2022}
  & ResNet-20 & \multicolumn{3}{c}{80.57 (T = 256)} \\
  \cmidrule(lr){2-6}
  & \multicolumn{5}{@{}l}{\textit{Direct SNN training}} \\
  & GLIF~\cite{yao2023glifunifiedgatedleaky}\venue{NeurIPS-2022}
  & ResNet-19 & -- & 71.87 & 72.57 \\
  & TET~\cite{deng2022temporalefficienttrainingspiking}\venue{ICLR-2022}
  & ResNet-19 & -- & 75.48 & 77.05 \\
  & RecDis-SNN~\cite{guo2022recdis}\venue{CVPR-2022}
  & ResNet-19 & -- & 76.64 & 78.53 \\
  & LT-SNN~\cite{hasssan2024spiking}\venue{IJCNN-2024}
  & ResNet-19 & -- & 72.78 & -- \\
  & Ternary Spike~\cite{guo_ternary_2024}\venue{AAAI-2024}
  & ResNet-20 & -- & 73.41 & 74.02 \\
  & DA-LIF~\cite{10888909}\venue{ICASSP-2025}
  & ResNet-20 & 69.37 & 72.07 & 73.25 \\
  & FSTA-SNN~\cite{yu2025fstasnnfrequencybasedspatialtemporalattentionmodule}\venue{AAAI-2025}
  & ResNet-20 & 69.64 & 72.15 & 73.44 \\
  & STAA-SNN~\cite{zhang2025staasnnspatialtemporalattentionaggregator}\venue{CVPR-2025}
  & ResNet-20 & 70.14 & 73.20 & 75.10 \\
  & LM-HT ($L=2$)$^{\dagger}$~\cite{hao2024lmht}\venue{NeurIPS-2024}
  & ResNet-20 & 78.74 & 79.96 & 80.67 \\
  \cmidrule(lr){2-6}
  & LBQ
  & ResNet-20 & \textbf{82.38} & \textbf{82.82} & \textbf{82.71} \\

  \bottomrule
  \end{tabularx}
  \vspace{2pt}

  \begin{minipage}{\textwidth}
  \footnotesize\textit{Note:} $^{\dagger}$LM-HT results are reproduced by us using ResNet-20. Other baseline results are taken from the cited papers. For ANN-to-SNN conversion methods, the actual simulation timestep is shown in parentheses.
  \end{minipage}
\end{table*}

\subsubsection{ImageNet-1K}
On ImageNet-1K, LBQ reaches 73.67\% with ResNet-34 at two timesteps, improving over the architecture- and timestep-matched LM-HT result by 2.77 percentage points (Table~\ref{tab:imagenet_eval}). This accuracy also exceeds the ResNet-34 Adaptive Calibration result at eight timesteps by 0.71 points. At one timestep, LBQ retains 72.45\% accuracy. ResNet-18 achieves 67.50\% and 68.23\% at one and two timesteps, respectively. Together, these results extend the value of learned burst resolution to large-scale image recognition.

\begin{table*}[t]
  \centering
  \caption{Classification accuracy (\%) on ImageNet-1K.}
  \label{tab:imagenet_eval}
  \footnotesize
  \setlength{\tabcolsep}{3pt}
  \renewcommand{\arraystretch}{1.05}
  \begin{tabularx}{\textwidth}{@{}lYlcc@{}}
  \toprule
  Type & Method & Architecture & Timestep & Acc. (\%) \\
  \midrule

  \multirow{4}{*}{\textbf{ANN-to-SNN}} 
  & Burst+LIPooling~\cite{li_efficient_2022}\venue{IJCAI-2022}
  & VGG-16 
  & 256 & 74.25 \\

  & Hybrid Training~\cite{rathi2020enablingdeepspikingneural}\venue{ICLR-2020}
  & ResNet-34
  & 250 & 61.48 \\

  & RMP-SNN~\cite{hanRmpsnnResidualMembrane2020}\venue{CVPR-2020}
  & ResNet-34
  & 4096 & 68.89 \\

  & Adaptive Calibration~\cite{Wang_Fang_Cao_Ren_Xu_2025}\venue{AAAI-2025}
  & ResNet-34
  & 8 & 72.96 \\
  \midrule

  \multirow{14}{*}{\textbf{SNN Training}}  
  & SEW~\cite{fangDeepResidualLearning2021}\venue{NeurIPS-2021}
  & ResNet-18 & 4 & 63.18 \\

  & Real Spike~\cite{guo2022realspikelearningrealvalued}\venue{ECCV-2022}
  & ResNet-18 & 4 & 63.68 \\

  & RMP-Loss~\cite{guo2023rmplossregularizingmembranepotential}\venue{ICCV-2023}
  & ResNet-18 & 4 & 63.03 \\

  & MPBN~\cite{guo2023membrane}\venue{ICCV-2023}
  & ResNet-18 & 4 & 63.14 \\
  & \multirow{2}{*}{LBQ}
  & \multirow{2}{*}{ResNet-18}
  & 1 & \textbf{67.50} \\

  & & & 2 & \textbf{68.23} \\
  \cmidrule(lr){2-5}
  & GLIF~\cite{yao2023glifunifiedgatedleaky}\venue{NeurIPS-2022}
  & ResNet-34 & 4 & 67.52 \\

  & RecDis-SNN~\cite{guo2022recdis}\venue{CVPR-2022}
  & ResNet-34 & 4 & 67.33 \\
  & LT-SNN~\cite{hasssan2024spiking}\venue{IJCNN-2024}
  & ResNet-34 & 4 & 68.46\\
  & Ternary Spike~\cite{guo_ternary_2024}\venue{AAAI-2024}
  & ResNet-34 & 4 & 70.74\\

  & FSTA-SNN~\cite{yu2025fstasnnfrequencybasedspatialtemporalattentionmodule}\venue{AAAI-2025}
  & ResNet-34 & 4 & 70.23 \\

  & LM-HT ($L=2$)~\cite{hao2024lmht}\venue{NeurIPS-2024}
  & ResNet-34 & 2 & 70.90 \\

  \cmidrule(lr){2-5}
  & \multirow{2}{*}{LBQ}
 & \multirow{2}{*}{ResNet-34}
  & 1 & \textbf{72.45} \\

  & & & 2 & \textbf{73.67} \\

  \bottomrule
  \end{tabularx}
\end{table*}

\subsection{Performance on Event-driven Datasets}
LBQ also supports event-driven recognition, reaching 80.50\% on CIFAR10-DVS with ResNet-20 at 10 timesteps and 98.28\% on DVS128-Gesture with ResNet-19 at 16 timesteps (Table~\ref{tab:dvs_eval}). On CIFAR10-DVS, it improves over our ResNet-20 LM-HT reproduction at the same timestep budget by 2.30 percentage points. It also matches the 80.50\% result reported for IMLIF with VGG-13 at 40 timesteps. These benchmarks demonstrate that the learned burst representation applies to both static images and temporally structured event inputs.

\begin{table*}[t]
  \centering
  \caption{Classification accuracy (\%) on event-driven datasets.}
  \label{tab:dvs_eval}
  \footnotesize
  \setlength{\tabcolsep}{3pt}
  \renewcommand{\arraystretch}{1.05}
  \begin{tabularx}{\textwidth}{@{}lYlcc@{}}
  \toprule
  Dataset & Method & Architecture & Timestep & Acc. (\%) \\
  \midrule

  \multirow{9}{*}{\textbf{CIFAR10-DVS}} 
  & SEW~\cite{fangDeepResidualLearning2021}\venue{NeurIPS-2021}
  & SEW-ResNet & 16 & 74.40 \\

  & GLIF~\cite{yao2023glifunifiedgatedleaky}\venue{NeurIPS-2022}
  & 7B-wideNet & 16 & 78.10 \\

  & RecDis-SNN~\cite{guo2022recdis}\venue{CVPR-2022}
  & ResNet-19 & 10 & 72.42 \\

  & MLF~\cite{feng2022multilevel}\venue{IJCAI-2022}
  & ResNet-14 & 10 & 70.36 \\

  & MPBN~\cite{guo2023membrane}\venue{ICCV-2023}
  & ResNet-20 & 10 & 78.70 \\

  & IMLIF~\cite{lian2024lif}\venue{TETCI-2024}
  & VGG-13 & 40 & \textbf{80.50} \\

  & LT-SNN~\cite{hasssan2024spiking}\venue{IJCNN-2024}
  & VGG-9 & 10 & 79.10 \\

  & Ternary Spike~\cite{guo_ternary_2024}\venue{AAAI-2024}
  & ResNet-20 & 10 & 79.80 \\

  & LM-HT ($L=2$)$^{\dagger}$~\cite{hao2024lmht}\venue{NeurIPS-2024}
  & ResNet-20 & 10 & 78.20 \\

  \cmidrule(lr){2-5}
  & LBQ
  & ResNet-20 & 10 & \textbf{80.50} \\

  \midrule
  \multirow{8}{*}{\textbf{DVS128-Gesture}} 
  & STBP-tdBN~\cite{zhengGoingDeeperDirectlytrained2021}\venue{AAAI-2021}
  & ResNet-17 & 40 & 96.87 \\

  & IMLIF~\cite{lian2024lif}\venue{TETCI-2024}
  & ResNet-19 & 40 & 97.33 \\

  & PLIF~\cite{fang2021incorporatinglearnablemembranetime}\venue{ICCV-2021}
  & PLIFNet & 20 & 97.57 \\

  & MLF~\cite{feng2022multilevel}\venue{IJCAI-2022}
  & ResNet-20 & 40 & 97.29 \\

  & MA-SNN~\cite{10032591}\venue{TPAMI-2023}
  & 5-layer SCNN & 20 & 98.23 \\

  & ASA-SNN~\cite{yao2023inherent}\venue{ICCV-2023}
  & 5-layer SCNN & 20 & 97.70 \\

  & SEW~\cite{fangDeepResidualLearning2021}\venue{NeurIPS-2021}
  & 7B-wideNet & 16 & 97.92 \\
  \cmidrule(lr){2-5}
  & LBQ
  & ResNet-19 & 16 & \textbf{98.28} \\
  \bottomrule
  \end{tabularx}
  \vspace{2pt}
  \begin{minipage}{\textwidth}
  \footnotesize\textit{Note:} $^{\dagger}$LM-HT results are reproduced by us using ResNet-20. Other baseline results are taken from the cited papers.
  \end{minipage}
\end{table*}

\begin{table*}[t]
  \centering
  \caption{Ablation study of LBQ $(N_{\max}=5)$ on CIFAR-10.}
  \label{tab:ablation}
  \footnotesize
  \setlength{\tabcolsep}{3pt}
  \renewcommand{\arraystretch}{1.08}
  \begin{tabularx}{0.98\textwidth}{@{}Ylccccc@{}}
  \toprule
  Method & Architecture & Spike output & Quant. step & Surrogate & Timestep & Acc. (\%) \\
  \midrule
  ANN & ResNet-20
  & Real-valued & N/A & N/A & 1 & 97.52 \\
  \midrule
  Binary LIF & ResNet-20
  & $\{0,1\}$ & N/A & Arctan & 2 & 91.81 \\

  Ternary Spike & ResNet-20
  & $\{-1,0,1\}$ & Fixed & Box-ET & 2 & 94.11 \\

  Ternary Spike & ResNet-20
  & $\{-1,0,1\}$ & Learnable & Box-ET & 2 & 95.55 \\

  V1 (Fixed + Box) & ResNet-20
  & $\{0,1,\ldots,N_{\max}\}$ & Fixed & Box-ET & 2 & 96.44 \\
  V2 (Fixed + ReLSG-ET) & ResNet-20
  & $\{0,1,\ldots,N_{\max}\}$ & Fixed & ReLSG-ET & 2 & 96.74 \\
  V3 (Learnable + Box) & ResNet-20
  & $\{0,1,\ldots,N_{\max}\}$ & Learnable & Box-ET & 2 & 97.30 \\

  LBQ (Learnable + ReLSG-ET) & ResNet-20
  & $\{0,1,\ldots,N_{\max}\}$ & Learnable & ReLSG-ET & 2 & \textbf{97.45} \\
  \bottomrule
  \end{tabularx}
\end{table*}

\subsection{Ablation Study}
\label{subsec:ablation}

LBQ reaches 97.45\% on CIFAR-10 at two timesteps, approaching the 97.52\% ANN reference with the same ResNet-20 architecture. Table~\ref{tab:ablation} identifies the contributions of burst representation, learned quantization, and surrogate training under common experimental settings. The four burst variants form a $2\times2$ comparison: V1 (Fixed + Box), V2 (Fixed + ReLSG-ET), V3 (Learnable + Box), and LBQ (Learnable + ReLSG-ET). Here, Fixed/Learnable specifies the quantization step, and Box denotes the Box-ET surrogate. We use V1--V3 as shorthand below.

\paragraph{Learning the burst resolution.}
Learning the quantization step improves accuracy under both surrogates. With Box-ET, accuracy rises from 96.44\% (V1) to 97.30\% (V3), a gain of 0.86 percentage points. With ReLSG-ET, it rises from 96.74\% (V2) to 97.45\% (LBQ), a gain of 0.71 points. The ternary variants show a corresponding gain from 94.11\% to 95.55\%. These controlled comparisons establish the benefit of optimizing the discretization step alongside synaptic weights.

\paragraph{Training across the burst range.}
ReLSG-ET improves the fixed-step model from 96.44\% to 96.74\% and the learned-step model from 97.30\% to 97.45\%. Its exponential tails extend gradient support beyond the active interval, complementing step learning in both settings. The full model combines these components and achieves the highest SNN accuracy in the ablation.

\paragraph{Expanding the discrete response space.}
Under the Box-ET surrogate, fixed-step burst responses reach 96.44\%, compared with 94.11\% for fixed-step ternary responses. This 2.33-point gain supports the use of multiple non-negative burst levels within each timestep. Learning their spacing then raises accuracy to 97.30\%, showing how adaptive resolution builds on the expanded response space.

\subsection{Layer-Wise Behavior of Learnable Burst Quantization}

\begin{figure*}[t]
  \centering
  \includegraphics[width=\textwidth]{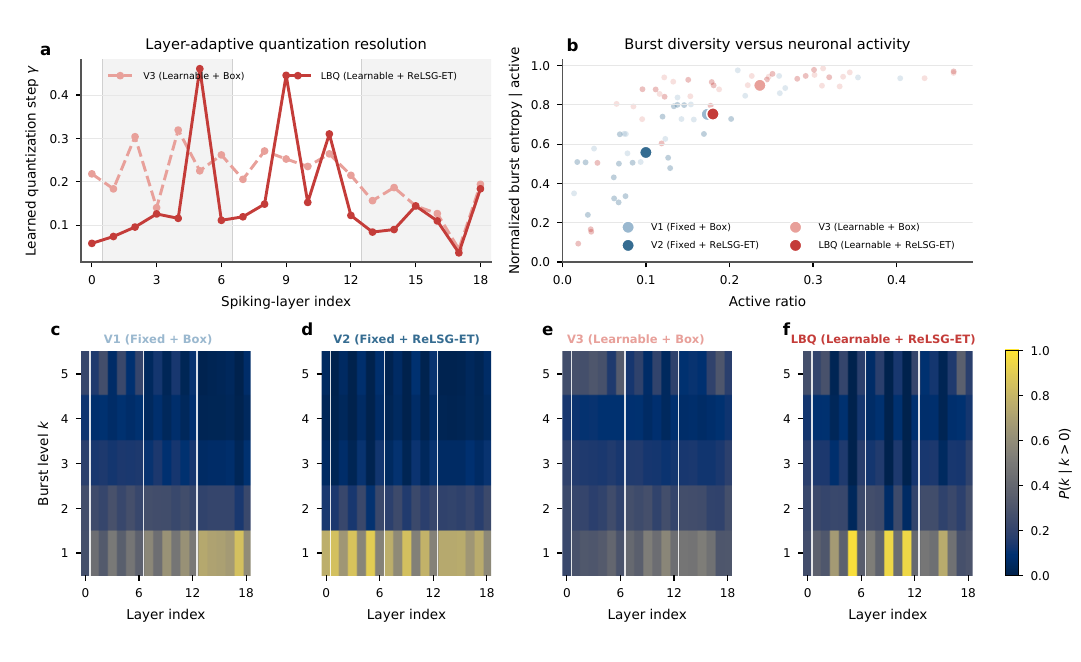}
  \caption{Layer-specific and selective burst allocation in trained ResNet-20 SNNs.
  (a) Learned quantization step $\gamma^{(\ell)}$ across spiking layers for V3 and the full LBQ model.
  (b) Layer-wise normalized conditional burst entropy versus neuronal active ratio; small markers represent individual layers and large markers represent model-wise means.
  (c--f) Conditional distributions $P(k\mid k>0)$ of non-zero burst levels across layers for V1 (Fixed + Box), V2 (Fixed + ReLSG-ET), V3 (Learnable + Box), and LBQ (Learnable + ReLSG-ET), respectively. Box denotes Box-ET.}
  \label{fig:layer_behavior}
\end{figure*}

\paragraph{Layer-specific resolution.}
Figure~\ref{fig:layer_behavior}(a) shows that full LBQ learns quantization steps spanning 0.036--0.461 across network depth. Its coefficient of variation is 0.74, compared with 0.31 for the learnable-step Box-ET model V3. The non-monotonic profile assigns different resolutions to individual layers, including pronounced maxima at intermediate layers and a minimum near the end of the network. This learned heterogeneity shows that LBQ does not collapse to a shared global resolution.

\paragraph{Selective burst responses.}
Full LBQ combines higher accuracy with a lower mean active ratio than V3: 97.45\% versus 97.30\% accuracy, with active ratios of 0.180 versus 0.236. Their normalized conditional burst entropies are 0.753 and 0.899, respectively (Fig.~\ref{fig:layer_behavior}(b)). The fixed-step comparison shows the same direction of change: replacing Box-ET with ReLSG-ET shifts the mean active ratio from 0.173 to 0.100 and normalized conditional entropy from 0.751 to 0.557. These profiles associate ReLSG-ET training with more selective firing across layers.

The conditional level distributions show how this selectivity is expressed (Fig.~\ref{fig:layer_behavior}(c--f)). The mean probability assigned to $k=1$ is 53.0\% for V1 and 69.7\% for V2. Learning the step broadens the use of higher levels in V3, whose mean conditional probability at $k=1$ is 38.2\%. Full LBQ assigns 49.8\% to $k=1$ while using levels $k=2$--5 in distinct patterns across depth. Combined with the ablation results, this pattern supports a representation in which each layer learns both its burst resolution and its use of the available levels.

\subsection{Bit-Sparse Accumulation and Arithmetic-Energy Analysis}
\label{subsec:energy}

\begin{table}[htbp]
  \centering
  \caption{Synaptic accumulation counts under unary and bit-sparse execution on CIFAR-10 with ResNet-20 at $T=2$.}
  \label{tab:bit_sparse_savings}
  \footnotesize
  \setlength{\tabcolsep}{3pt}
  \renewcommand{\arraystretch}{1.08}
  \begin{tabularx}{\columnwidth}{@{}Yrrr@{}}
    \toprule
    Method & \makecell{Unary-equiv.\\ACs (M)} & \makecell{Bit-sparse\\ACs (M)} & \makecell{Reduction\\(\%)} \\
    \midrule
    Binary & 79.59 & 79.59 & 0.00 \\
    Ternary & 1693.84 & 1693.84 & 0.00 \\
    \midrule
    LBQ (2) & 727.58 & 542.72 & 25.41 \\
    LBQ (5) & 1589.90 & 945.69 & 40.52 \\
    LBQ (10) & 2577.73 & 1203.80 & 53.30 \\
    LBQ (20) & 2828.76 & 1163.20 & \textbf{58.88} \\
    \bottomrule
  \end{tabularx}
  \vspace{2pt}

  \begin{minipage}{\columnwidth}
  \footnotesize\textit{Note:} LBQ ($N_{\max}$) specifies burst capacity. For LBQ, unary-equivalent and bit-sparse ACs are $C_{\mathrm{AC}}^{\mathrm{unary}}$ and $C_{\mathrm{AC}}^{\mathrm{BS}}$ in Eqs.~\eqref{eq:unary_acs} and~\eqref{eq:bit_sparse_acs}; Reduction is $100(1-C_{\mathrm{AC}}^{\mathrm{BS}}/C_{\mathrm{AC}}^{\mathrm{unary}})$. For Binary/Ternary, both columns report conventional unit-event ACs. Formation and explicit shift operations are excluded.
  \end{minipage}
\end{table}

\begin{table*}[t]
  \centering
  \caption{Accuracy and modeled arithmetic-cost comparison on CIFAR-10 using ResNet-20; all SNN models use $T=2$.}
  \label{tab:energy}
  \footnotesize
  \setlength{\tabcolsep}{1.5pt}
  \renewcommand{\arraystretch}{1.05}
  \begin{tabularx}{\textwidth}{@{}Yrrrrrr@{}}
      \toprule
      Method & ACC (\%) & \makecell{FP MACs$^{\dagger}$\\(M)} & \makecell{Synaptic\\ACs (M)} & \makecell{Bit-Plane Formation\\Ops (M)} & \makecell{Modeled energy\\(mJ)} & \makecell{Modeled energy ratio\\$E_{\mathrm{ANN}}/E$} \\
      \midrule
      ANN & 97.52 & 2587.24 & N/A & N/A & 11.90 & -- \\
      \midrule
      Binary & 91.81 & 3.54 & \textbf{79.59} & N/A & \textbf{0.09} & \textbf{135.38} \\
      Ternary & 94.11 & 3.54 & 1693.84 & N/A & 1.54 & 7.72 \\
      LBQ (2) & 96.73 & 3.54 & 542.72 & 0.385 & 0.50 & 23.58 \\
      LBQ (5) & \textbf{97.45} & 3.54 & 945.69 & 0.695 & 0.87 & 13.72 \\
      LBQ (10) & 97.44 & 3.54 & 1203.80 & 0.938 & 1.10 & 10.82 \\
      LBQ (20) & 97.35 & 3.54 & 1163.20 & 0.983 & 1.06 & 11.19 \\
      \bottomrule
  \end{tabularx}
  \vspace{2pt}

  \begin{minipage}{\textwidth}
  \footnotesize\textit{Note:} LBQ ($N_{\max}$) specifies the maximum burst level. $^{\dagger}$FP MACs cover the full ANN and the first layer of each SNN. Synaptic ACs count unit-event accumulations for Binary/Ternary and $C_{\mathrm{AC}}^{\mathrm{BS}}$ for LBQ [Eq.~\eqref{eq:bit_sparse_acs}]. Bit-Plane Formation Ops count $C_{\mathrm{BPF}}$ before synaptic fan-out. The modeled energy assumes pre-shifted weights or fused power-of-two scaling; explicit runtime synaptic shifts are excluded. Ratios use unrounded estimates.

  \end{minipage}
\end{table*}

\paragraph{Bit sparsity reduces burst accumulation cost.}
Table~\ref{tab:bit_sparse_savings} isolates the effect of bit-sparse execution by comparing unary-equivalent and bit-sparse synaptic ACs for the same trained models. At $N_{\max}=5$, the accumulation count decreases from 1589.90 to 945.69 million, a saving of 40.52\%. Across the four LBQ configurations, the reduction ranges from 25.41\% to 58.88\%, with $N_{\max}=20$ reducing the count from 2828.76 to 1163.20 million. Binary and unit-magnitude ternary events retain their conventional event-AC counts. The LBQ savings follow from representing each burst level $S$ by its $\operatorname{popcount}(S)$ non-zero bits. Across trained models, the accumulation count depends on the learned level distribution as well as the configured burst capacity. Bit-plane formation is accounted for separately in Table~\ref{tab:energy}.

\paragraph{Accuracy and arithmetic cost.}
Table~\ref{tab:energy} separates floating-point MACs, synaptic ACs, and bit-plane formation operations. We estimate arithmetic energy using 4.6\,pJ per 32-bit FP MAC and 0.9\,pJ per accumulation from the 45-nm CMOS model of Horowitz~\cite{horowitz2014computing}, following the MAC/accumulation convention in Vision SmolMamba~\cite{bai2026visionsmolmamba}. We use 0.1\,pJ, the 32-bit integer-add cost, as a proxy for each abstract bit-plane formation operation:
\begin{equation}
\label{eq:energy_proxy}
\begin{aligned}
E_{\mathrm{arith}}
={}&4.6\,\mathrm{pJ}\,C_{\mathrm{FP\text{-}MAC}}
+0.9\,\mathrm{pJ}\,C_{\mathrm{AC}}
\\
&+0.1\,\mathrm{pJ}\,C_{\mathrm{BPF}}.
\end{aligned}
\end{equation}
For LBQ, $C_{\mathrm{AC}}=C_{\mathrm{AC}}^{\mathrm{BS}}$. The reported proxy corresponds to a deployment in which power-of-two scaling is precomputed or fused with accumulation. A backend that issues explicit runtime shifts adds $E_{\mathrm{shift}}C_{\mathrm{shift}}^{\mathrm{syn}}$ to Eq.~\eqref{eq:energy_proxy}, using the count in Eq.~\eqref{eq:explicit_shifts}. If one reported formation event requires $q$ integer ALU instructions, its contribution becomes $0.1\,\mathrm{pJ}\,qC_{\mathrm{BPF}}$. Memory traffic, routing, and neuron-state updates remain separate from this arithmetic model.

LBQ ($N_{\max}=5$) combines 97.45\% accuracy with an estimated arithmetic cost of 0.87\,mJ, compared with 97.52\% and 11.90\,mJ for the ANN reference. Under the reported AC-plus-formation proxy, this corresponds to a $13.72\times$ reduction in modeled arithmetic energy at near-ANN accuracy. A smaller capacity, $N_{\max}=2$, reaches 96.73\% at 0.50\,mJ, providing a lower-cost operating point. The two settings illustrate how burst capacity controls the accuracy--arithmetic-cost trade-off while bit-plane execution exploits the integer representation in both cases.

\section{Conclusion}

Learnable Burst Quantization (LBQ) recasts burst spiking as saturated uniform quantization with a layer-wise learned membrane-to-burst step. ReLSG-ET supports joint optimization of burst resolution and synaptic weights by sustaining gradient support throughout and beyond the active burst range. At inference, quantization-step absorption preserves the scaled synaptic input, while integer-level bit-plane decomposition enables bit-sparse synaptic accumulation over only the non-zero planes. Experiments across static and event-driven benchmarks establish the broad applicability of this formulation. Controlled ablations isolate the roles of learned quantization and ReLSG-ET, while layer-wise analyses reveal selective burst allocation across network depth. On CIFAR-10 at $T=2$, LBQ reaches 97.45\%, only 0.07 percentage points below the 97.52\% architecture-matched ResNet-20 ANN reference. In the same setting, bit-plane execution at $N_{\max}=5$ reduces unary-equivalent synaptic accumulations by 40.52\% relative to unary execution. Together, these results establish adaptive burst quantization as an effective route to accurate few-timestep SNN inference with an algebraically equivalent bit-sparse execution path.

\section*{Data and Code Availability}
The implementation code is available at
\url{https://https://github.com/dwbai95/LBQ}. The datasets used in this study are publicly available benchmarks, as described in the experimental setup.

\bibliographystyle{unsrtnat}
\bibliography{references}

\end{document}